\documentclass[letterpaper, 10 pt, conference]{ieeeconf}  % Comment this line out if you need a4paper

\IEEEoverridecommandlockouts                              % This command is only needed if 
\usepackage{amsmath} % assumes amsmath package installed

\usepackage[
  colorlinks,
  breaklinks,
  pdftitle={},
  pdfauthor={ITU Artificial Intelligence and Data Science Application and Research Center},
  unicode
]{hyperref}
\usepackage[linesnumbered,ruled,vlined]{algorithm2e}
\usepackage{svg}
\usepackage{mathrsfs}
\usepackage{subcaption}
\SetCommentSty{mycommfont}

\SetKwInput{KwInput}{Input}                % Set the Input
\SetKwInput{KwOutput}{Output}              % set the Output
\usepackage[noend]{algpseudocode}
\usepackage{graphicx}
\usepackage{float}
\usepackage{siunitx}

\title{\LARGE \bf Investigating Value of Curriculum Reinforcement Learning in Autonomous Driving Under Diverse Road and Weather Conditions}

\author{Anil Ozturk$^{\ast}$,  Mustafa Burak Gunel$^{\ast}$, Resul Dagdanov, Mirac Ekim Vural, Ferhat Yurdakul, Melih Dal  \\
and Nazim Kemal Ure % <-this % stops a space

\thanks{$^\ast$ These authors contributed equally to this work}% <-this % stops a space
\thanks{A. Ozturk is with ITU Artificial Intelligence and Data Science Research Center and Department of Computer Engineering, Istanbul Technical University, Turkey 
        {\tt\small ozturka18 at itu.edu.tr}}%
\thanks{M.B. Gunel is with ITU Artificial Intelligence and Data Science Research Center and Department of Aeronautical Engineering, Istanbul Technical University, Turkey
        {\tt\small mustafa.gunel at itu.edu.tr}}%
\thanks{R. Dagdanov is with ITU Artificial Intelligence and Data Science Research Center and Department of Aeronautical Engineering, Istanbul Technical University, Turkey
        {\tt\small dagdanov17 at itu.edu.tr}}%
\thanks{M.E. Vural is with ITU Artificial Intelligence and Data Science Research Center and Department of Mechanical Engineering, Istanbul Technical University, Turkey
        {\tt\small vuralm15 at itu.edu.tr}}%
\thanks{F. Yurdakul is with ITU Artificial Intelligence and Data Science Research Center and Department of Aeronautical Engineering, Istanbul Technical University, Turkey
        {\tt\small yurdakul17 at itu.edu.tr}}%
\thanks{M. Dal is with Faculty of Computer Engineering, Bogazici University, Turkey 
        {\tt\small melih.dal at boun.edu.tr}}%
\thanks{N.K. Ure is with ITU Artificial Intelligence and Data Science Research Center and Department of Aeronautical Engineering, Istanbul Technical University, Turkey
        {\tt\small ure at itu.edu.tr}}%
}

\begin{document}
\maketitle
\thispagestyle{empty}
\pagestyle{empty}

%%%%%%%%%%%%%%%%%%%%%%%%%%%%%%%%%%%%%%%%%%%%%%%%%%%%%%%%%%%%%%%%%%%%%%%%%%%%%%%%%%%%%%%%%%%%%%%%%%%%%%%
\begin{abstract}
Applications of reinforcement learning (RL) are popular in autonomous driving tasks. That being said, tuning the performance of an RL agent and guaranteeing the generalization performance across variety of different driving scenarios is still largely an open problem. In particular, getting good performance on complex road and weather conditions require exhaustive tuning and computation time. Curriculum RL, which focuses on solving simpler automation tasks in order to transfer knowledge to complex tasks, is attracting attention in RL community. The main contribution of this paper is a systematic study for investigating the value of curriculum reinforcement learning in autonomous driving applications. For this purpose, we setup several different driving scenarios in a realistic driving simulator, with varying road complexity and weather conditions. Next, we train and evaluate performance of RL agents on different sequences of task combinations and curricula. Results show that curriculum RL can yield significant gains in complex driving tasks, both in terms of driving performance and sample complexity. Results also demonstrate that different curricula might enable different benefits, which hints future research directions for automated curriculum training.  
% Reinforcement Learning (RL) is a highly fragile progress that depends on correct selection of hyper-parameters. Solving complex RL problems in complex environments such as real-world like traffic scenarios is extremely hard. Curriculum reinforcement learning helps and stabilizes this training process. It can also improve the performance of the trained agents significantly. There are some examples of this approach in autonomous driving, but there are not many studies that try to solve this problem systematically in a realistic traffic simulator. In this work, a proprietary traffic simulator developed by Eatron Technologies has been been used. It can simulate realistic vehicle dynamics and weather conditions. Different custom road types have been created and an RL agent has been trained in each of these scenarios with different weather conditions that affects the road dynamics. It has been demonstrated through simulations that the use of curriculum learning vastly improved the performance of the agents and the training times in relatively complex environments.\kXX{Check}
\end{abstract}
%%%%%%%%%%%%%%%%%%%%%%%%%%%%%%%%%%%%%%%%%%%%%%%%%%%%%%%%%%%%%%%%%%%%%%%%%%%%%%%%%%%%%%%%%%%%%%%%%%%%%%%
\section{Introduction} \label{section:introduction}
There has been a huge improvement in autonomous driving systems in the recent years, thanks to the development of novel machine learning algorithms and increasing processing power of computer hardware. In particular, training autonomous driving agents in realistic simulators has been a popular choice for developing reinforcement learning (RL) algorithms \cite{bicer}\cite{iv20}. Although it is possible to simulate and train agents on simple environments, training on a complex and realistic simulator is still an open problem due to the development issues and computational costs. That being said, there is also a great value in creating agents that can operate well across a variety of different weather and road conditions. The USDOT Highway Administration data \cite{ref16} shows that the $22 \%$ of all vehicle accidents are due to the adverse weather conditions, which highlights the significance of this problem. The main promise of autonomous driving is that it can decrease accidents and fatalities, rising from difficult driving conditions, such as adverse weather conditions. Therefore, the problem of training an agent that is robust to diversity of adverse weather and road conditions is an important challenge. The standard way to tackle this problem in the context of RL is developing complex simulation environments that corresponds to the most challenging situations and train agents on them. However, as the complexity of the environment increases, RL agents struggle to discover high quality solutions, which leads to overly long training times and poor driving performance. In recent years, curriculum RL methodologies \cite{narvekar} started to attract attention to solve such complicated decision making tasks. The main idea behind curriculum RL is, instead of tackling the complex decision making task directly, solving similar but simpler problems first and then transferring this knowledge to the complex problem might yield better returns.

\smallskip

Even though curriculum RL was used in autonomous driving in previous work (will be mentioned in Section \ref{subsection:prev}), there has been no in-depth study to investigate the benefits of this approach in a realistic driving simulator under different road and weather conditions. The main contribution of this work is a systematic study and comparison of different curricula on training RL agents on complex and realistic driving scenarios. Our results show that choosing the appropriate curriculum can significantly boost the driving performance and enable convergence with less samples from the simulator. 

% Executing this comprehensive training in an optimized manner also has an importance. The problem with using complex simulators and traffic conditions is the time for a reinforcement agent to learn the underlying dynamics of the environment increases significantly. In order to tackle with this problem a curriculum learning approach have been proposed to both increase the performance of the agents in different simulation conditions and decrease the overall training time to achieve those results.
\subsection{Previous Work}
\label{subsection:prev}
Curriculum learning is a meta-learning methodology that starts learning how to solve a problem from simpler examples and gradually increases the complexity of the examples in different thresholds. One of the earlier successful examples of using this methodology in machine learning is the Projection Problem study by Elman \cite{ref1} and observation of the effects of curriculum learning on training time and effectiveness by Bengio \cite{bengio}. Curriculum methods has also been adapted to reinforcement learning. The usage of curriculum learning in recent works \cite{ref2}, \cite{ref3} and Teacher-Student Curriculum Learning Framework \cite{ref4} shows that the curriculum learning improved performance of the training. The usage of another curriculum learning framework Mix \& Match \cite{ref5} shows that this approach can also decrease the sample complexity of the training process. There are also various successful applications of curriculum learning in other fields like robotics \cite{ref7}, language modelling \cite{ref8} and motion planning \cite{ref9}. A different study \cite{ref11} proposes an extension called SafeDAgger to DAgger \cite{ref10}, which is an advanced imitation learning algorithm. TORCS \cite{TORCS} racing simulator was used for evaluating the proposed method. It has been observed that the proposed approach achieves a better performance compared with the DAgger method and requires less queries. Another study \cite{ref13} uses Deep Reinforcement Learning to learn driving behavior in urban intersections. From a sample set of tasks, the authors automatically generated curricula for training phase according to their total rewards obtained from samples. After obtaining the curricula, an RL algorithm like DQN has been used to train agents. It has been observed that compared with the randomly generated sequences, using automatically generated curricula  significantly reduces the sample complexity of training. The usage of curriculum learning in smooth maneuvering on highways \cite{ref14} also yields good results. The study shows promising performance on overtaking and the decreasing of training times. However, being trained on simple TORCS environment hurts its ability to be implemented in real life use cases.  

\smallskip

To summarize, the usage of curriculum reinforcement learning in a variety of different autonomous driving tasks shows great potential in improving learning performance and convergence rate. That being said, there is a lack of in-depth study for comparing the impact of different curricula on the driving performance, especially in realistic driving simulators that can capture the real-world dynamics.

% different and complex tasks has decreased the overall training time and improve the performance of the trained agents. They achieved great results, but they have generally been trained in simple environments.

\subsection{Contributions} \label{subsection:contribution}

The main contribution of this work is an in-depth study on impact of different curricula on deep reinforcement learning for autonomous driving in a realistic driving simulator. We develop a structured environment, where the adversity of weather conditions and road complexity can be tuned independent from each other. We setup several different curricula, where the training starts from simple weather conditions and road geometries, and then ramps up to more complex road and weather conditions. Evaluation results show that agents trained using curriculum reaches superior performance using much lesser samples, compared to agents that are directly trained in complex environments.

\section{Background}
%%%%%%%%%%%%%%%%%%%%%%%%%%%%%%%%%%%%%%%%%%%%%%%%%%%%%%%%%%%%%%%%%%%%%%%%%%%%%%%%%%%%%%%%%%%%%%%%%%%%%%%
\subsection{Reinforcement Learning} 
\label{subsection:RL} In this section, the main ideas in RL are reviewed. The reader is referred to \cite{sutton}, for an in-depth discussion of RL. In reinforcement learning, agents take actions in an environment to maximize their cumulative rewards in an episode that consists of finite number of steps. The agents' actions at any state is defined in terms of a policy. Agents tune their policies with the penalties or rewards they get. There are two kinds of reinforcement learning algorithm in terms of how the policies are used; on-policy and off-policy.

\smallskip

%%%%%%%%%%%%%%%%%%%%%%%%%%%%%%%%%%%%%%%%%%%%%%%%%%%%%%%%%%%%%%%%%%%%%%%%%%%%%%%%%%%%%%%%%%%%%%%%%%%%%%%
In on-policy reinforcement learning, the policy used for updating the target model and the policy used for selecting the actions (behavior model) are the same. In off-policy reinforcement learning, the policy used for updating the target model and the policy used for selecting the actions (behavior model) can be different. Updating the model only requires specific inputs like state, action, next state and reward. In off-policy RL, the target policy may be deterministic, while the behavior policy can act on the sampled states from all possible past scenarios. This flexibility enables training the algorithm using previous experiences, instead of being constrained by the most up to date examples collected from the environment. In this context, all collected iteration data are stored in one place. In each training iteration, a certain amount of data is sampled from this collected data. This set where we sample the experience is called the buffer, and the sampled data is called batch.

\smallskip

%%%%%%%%%%%%%%%%%%%%%%%%%%%%%%%%%%%%%%%%%%%%%%%%%%%%%%%%%%%%%%%%%%%%%%%%%%%%%%%%%%%%%%%%%%%%%%%%%%%%%%%
As mentioned before, the aim in reinforcement learning is computing a policy that yields optimal actions for the agent. Policy gradient methods are based on updating the policy directly. An off-policy method learns the values of the optimal actions regardless of the agent's current actions. Policy is usually defined through a parameterized function, like a neural network. The generalized reward function for an off-policy learner is defined as:
\begin{align}
    J(\theta) = \sum_{s \in S}d^{\beta}(s) \sum_{a \in A}\pi_{\theta}(a|s)Q^{\pi}(s,a) \label{pg_eq}
\end{align}
In Eq. \ref{pg_eq}, $d^{\beta}(s)$ is the distribution of the behavior policy $\beta$. Without using the behavior policy, the value function $Q^{\pi}$ can be calculated only by the target policy. This difference is what makes an algorithm off-policy. The $Q$ function can be structured as:
\begin{align}
    Q_{\pi}(s,a) = E_{\pi}\{R_d|s_t=s, a_t=a\} \label{q_eq}
\end{align}
$Q_{\pi}(s,a)$ is action-value function for policy $\pi$. It returns the expected total discounted reward $R_d$ from the time-step $t$, for starting from state $s$, taking the action $a$ with policy $\pi$.
%%%%%%%%%%%%%%%%%%%%%%%%%%%%%%%%%%%%%%%%%%%%%%%%%%%%%%%%%%%%%%%%%%%%%%%%%%%%%%%%%%%%%%%%%%%%%%%%%%%%%%%

\smallskip

The Deterministic Policy Gradient (DDPG) \cite{ddpg} is one of the popular off-policy algorithms that learns the $Q$ function. Soft Actor Critic (SAC) \cite{sac} is an off-policy algorithm that uses the benefits of both stochastic policy optimization and DDPG approaches. The main part of the SAC is entropy regularization. The policy aims to increase the relationship between expected return and entropy as much as possible. The exploration-exploitation dilemma is directly related with the entropy. In this study, reinforcement models were trained in various highway scenarios using the SAC architecture.

\section{Simulation Enviroment}%SimStar\texttrademark \ Environment}
In this section, the simulation environment SimStar\footnote{ developed by Eatron Technologies Ltd., UK} is introduced. All training and evaluation procedures applied in this work are implemented in this simulation environment. In the environment, custom roads with various traffic, track, and weather conditions can be generated easily for more realistic-comprehensive training and evaluation of RL agents. This engine is responsible for 3D visualization of the environment and creating accurate vehicle dynamics. Different types of vehicles (sedan, SUV, truck etc.) can be added to the environment. Road conditions such as tar, dirt, damage can be implemented on the tracks to accurately resemble real-world counterparts. 

\subsection{Vehicle Dynamics}
SimStar makes use of NVIDIA PhysX \cite{noauthor_vehicles_nodate} as the primary physics engine. It is a highly sophisticated physics engine, and the realism in vehicle dynamics validated  through various  real-life  experiments. Additionally, the vehicle properties are adjusted to match a real autonomous vehicle. They are also validated through real-life road tests. 

\subsection{Weather Effects on Physics}
Accurate simulation of weather effects depends on modelling of the interaction between the tire and the road. Pajecka Tire Model \cite{pacejka_chapter_2012} is used as the tire model. The parameters regarding road friction and tire friction are calculated to match a real world study on the topic. The work by Kordani et. al \cite{kordani_effect_2018} calculates the road friction coefficient at different weather conditions for different type of vehicles. 

\begin{table}[H]
\centering 
\caption{Braking Distance of Vehicles on Adverse Conditions} 
\begin{tabular}{llll}
 & Dry & Rainy  & Snowy  \\
 \hline  \hline
Coefficient of Friction & 0.8 & 0.4 & 0.28  \\
Sedan  (m) & 105 & 114 & 133  \\
Bus  (m) & 115 & 116 & 169 \\ 
Sedan  (m) (SimStar) & 80.5 & 84.1 & 91.0  \\
 \hline
\end{tabular}
\label{table:frompaper}
\end{table}

The results for the difference in braking differences on adverse weather conditions is used in this work's simulations to generate realistic behavior. Braking distance of a vehicle going at $80 kph$ can be seen at Table \ref{table:frompaper} at each adverse condition. Since the vehicles used in the simulator are different than the vehicles in the reference paper, only modeling the adverse weather effect solutions are opted in proportion to the original study. The breaking distance of carefully modeled \textit{sedan} vehicle on dry weather is $80.5$ meters. Then, the rest of the adverse weather road models are adapted to create the correct proportional effect on braking distances. The final results can also be seen on the Table \ref{table:frompaper}.

\subsection{Observation and Control}
 All vehicles get several information about the road and the vehicle itself as well as information about other vehicles at every control step. SimStar provides several sensors on every vehicle in order to supply the information demands of the control algorithms. The road deviation sensor gives these information about the vehicle: 
 \begin{itemize}
     \item The vehicle’s angular deviation from the road’s central axis in radians.
     \item The vehicle’s distance deviation from the road’s central axis in meters.
 \end{itemize}

 These two values are scalars and are included in the observation at every control step. 

\iffalse 
\begin{figure}[H]
    \centering
    \includegraphics[width=0.395\textwidth]{figures/opponent.png}
    \caption{The laser sensor which returns the distance of each ray from vehicle to road boundaries.}
    \label{fig:sensor}
\end{figure} 
 \fi
 
 The track sensor receives information about the vehicle’s location on the road. This sensor is used to identify borders and edges of the road with respect to the vehicle's central body. The sensor gives a vector of $19$ scalar values. These values are being created by scanning the front-half of the surrounding area with $10^{\circ}$ splits. Thus, there are $19$ distance sensor lines. These values are also included in the observation state in the environment at every action step.

\smallskip

%%%%%%%%%%%%%%%%%%%%%%%%%%%%%%%%%%%%%%%%%%%%%%%%%%%%%%%%%%%%%%%%%%%%%%%%%%%%%%%%%%%%%%%%%%%%%%%%%%%%%%%
\section{Experimental Setup}
\subsection{Curriculum Setup}
We describe the details regarding curriculum implementation in this section. All agents are trained with the SAC algorithm in different scenarios and conditions. In this context, the agent is firstly trained exclusively in different weather conditions, which are: "clear", "rainy" and "snowy". 

\begin{figure}[H]
    %\centering
    
    \begin{minipage}{.27\textwidth}
        \begin{subfigure}{\textwidth}
        \centering
        \fbox{\includegraphics[width=\textwidth]{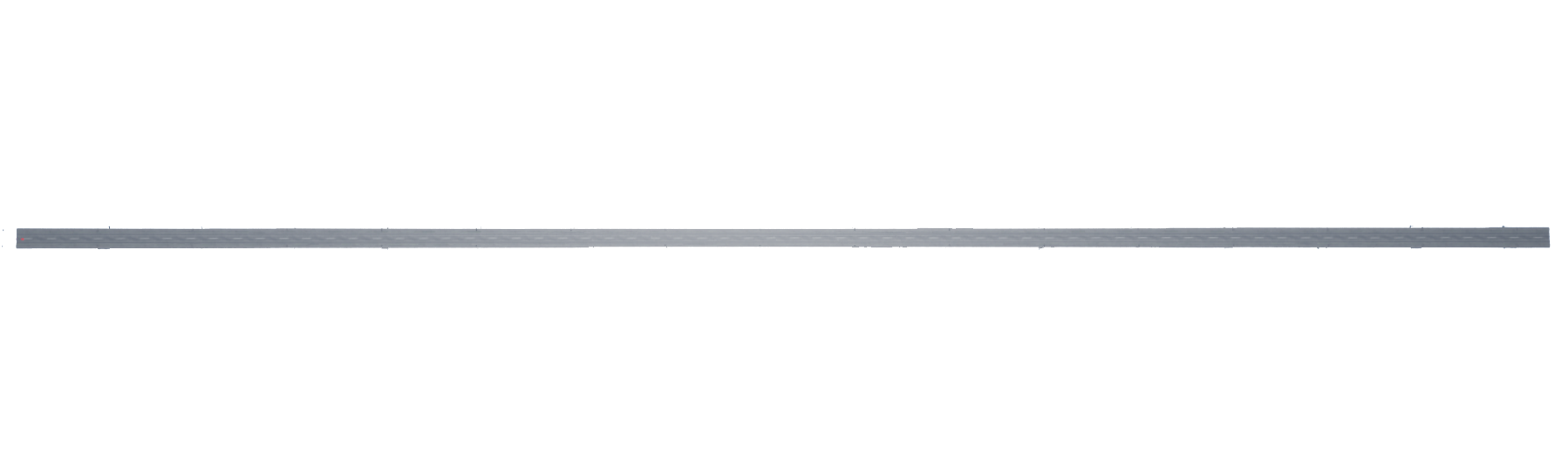}}
        \end{subfigure}
       \vspace{0.22cm}

     \end{minipage}
     \begin{minipage}{.05\textwidth}
     \smallskip
     \end{minipage}
    \begin{minipage}{.175\textwidth}
        \begin{subfigure}{\textwidth}
        \centering
        \fbox{\includegraphics[width=\textwidth]{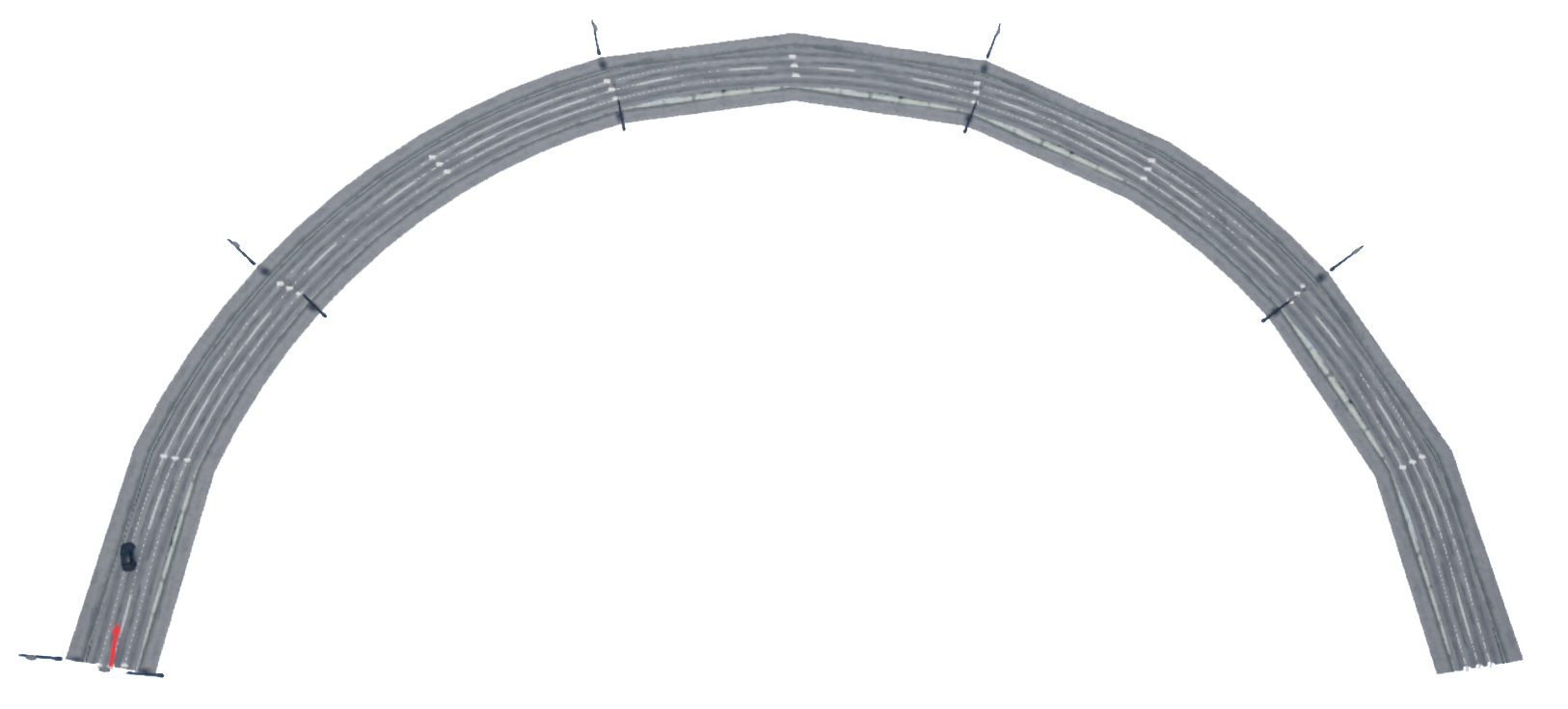}}
        \end{subfigure}
        \vspace{0.22cm}

     \end{minipage}
     \begin{minipage}{.05\textwidth}
     \smallskip
     \end{minipage}
     \fbox{\includegraphics[width=.975\linewidth]{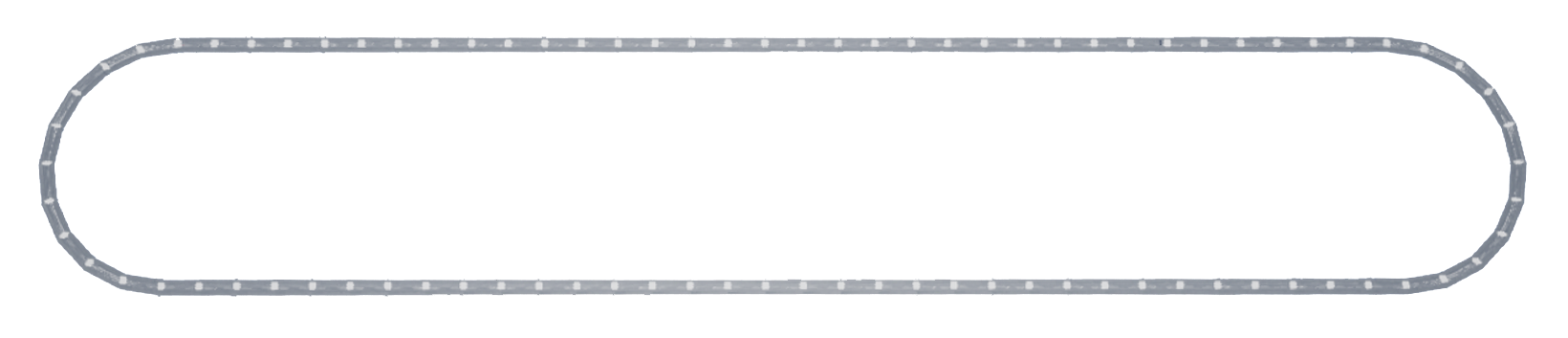}}     
     \caption{Road geometries used in simulation setup. (The Straight Road at the upper-left, the U-Turn Road at the upper-right and Complete Circuit Track at the bottom.}
     \label{fig:road_types}
\end{figure}

Another curriculum parameter is the geometry complexity of the track. The track that has the most complex geometry is the hardest track to be learned by the agent. All roads in the simulation are $10$ meters wide and consist of $2$ different lanes. Three tracks in total are used. As it can be seen on Figure \ref{fig:road_types}, they are;

\smallskip

\begin{itemize}
    \item Straight Road
    \item U-Turn Road
    \item Complete Circuit Race Track
\end{itemize}

\smallskip

To increase the variety of the curriculum learning, weather conditions are added onto these tracks for more realistic settings. The agents which are trained on different weather conditions (such as rain and snow) would learn unique driving capabilities. Note that clear, rainy, and snowy weather conditions are chosen in their maximum value for noteworthy atmospheric ambiances and their effects on the tracks. As a consequence, the curriculum reinforcement learning methodology is implemented on "Weather Condition" and "Track Type" combinations.

\smallskip

\begin{figure}[H]
    \centering
    \includegraphics[width=0.40\textwidth]{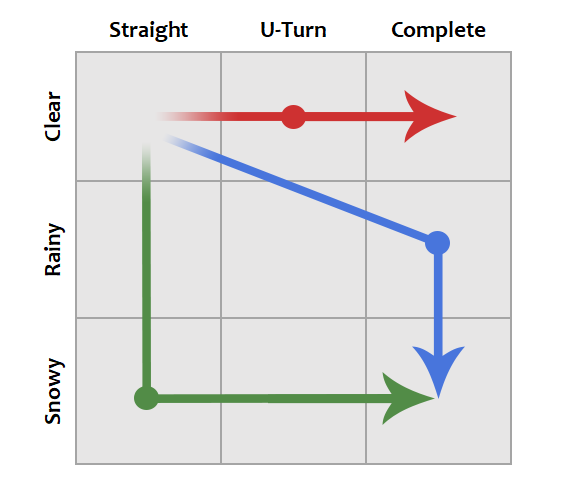}
    \caption{Three example hand-made curriculum scenarios with different road type and weather condition combinations. Each dot represent a combination change in the curriculum. Arrow tips represent the final state.}
     \label{fig:curri_grid}
\end{figure}

Each variable set of curriculum training scenarios can be evaluated as separate dimensions in a space. The transition process after each curriculum training phase will take place in the form of a transition from one point in the space to another. Figure \ref{fig:curri_grid} can be examined as a representation of the mentioned space and transitions.

\smallskip

\begin{figure}[H]
    \centering
    \includegraphics[width=0.42\textwidth]{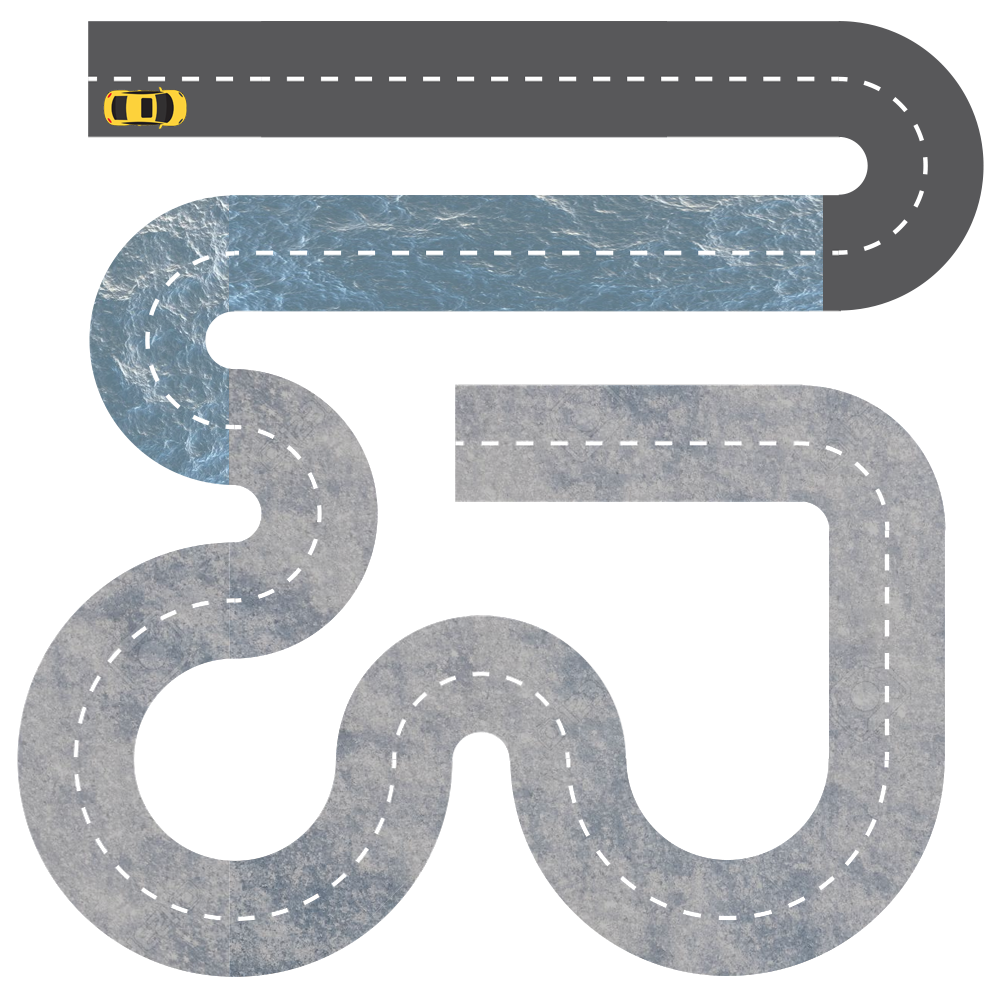}
    \caption{Representation of a full curriculum training scenario, which consists of straight, u-turn, complete roads and clear, rainy, snowy weather conditions, respectively.}
    \label{fig:curri_sc}
\end{figure}

An example and a complete curriculum scenario can be seen on Figure \ref{fig:curri_sc}. The agent observes the phases (Straight, Clean), (U-turn, Clean), (Straight, Rainy), (U-turn, Rainy), (Complete, Snowy) respectively. In the example design, the agent is trained only $1$ iteration on each phase and transferred to the other phase immediately. These iteration numbers will be in thousands in normal applications and the agents can be thought of as spawning back to the beginning of the current phase at the end of each iteration. When the iteration number of each phase is exceeded, the transition to the next phase will be executed.

\begin{table}[H]
\caption{Training Combinations}
\begin{tabular}{|c|c|c|c|}
\hline
\textbf{\begin{tabular}[c]{@{}c@{}}Curriculum \\ Scenarios\end{tabular}} & \textit{Phase 1} & \textit{Phase 2}   & \textit{Phase 3}   \\\hline
\textit{Scenario 1}               & Straight (C) & U-Turn (C)     & Race Track (C) \\\hline
\textit{Scenario 2}               & U-Turn (R)   & Race Track (R) & Race Track (C) \\\hline
\textit{Scenario 3}                & Straight (R) & Race Track (R) & Race Track (S) \\\hline
\textit{Scenario 4}                 & Straight (C) & Race Track (R) & Race Track (S) \\\hline
\textit{Scenario 5}                  & Straight (C) & Straight (S)   & Race Track (S) \\\hline
\end{tabular}
\label{table:tra_combs}
\end{table}

The complete information about training scenarios can be found in Table \ref{table:tra_combs}. A total of 5 different curriculum scenarios are implemented. Letters C, R and S corresponds to clear, rainy and snowy weather conditions respectively. These route orders are determined by examining the training results without curriculum learning. Most of the cases, the routes are chosen to be ordered from easiest scenarios to hardest. The order of curriculum learning is crucial as the agent has to acquire useful experience in the consecutive tasks \cite{narvekar}. It is seen that, the hardest scenario for the RL agent to solve is complete track scenario with snowy weather condition. So, the goal is to train the agent starting from the easiest track to be learned to the hardest track. 

\iffalse
To do so, specific training orders in Table \ref{table:tra_combs} are determined where agent starts from the simple environment and finishes at more complex one. By doing so, the agent will gather convenient experiences throughout the ordered training tracks.
\fi

\smallskip

The curriculum training order for each route is illustrated with ascending enumeration. The detailed training scenarios will be discussed in Section \ref{sec:results}. As stated in \cite{narvekar}, the quality of the curriculum learning task is directly dependent on the training order quality of the tracks.

\subsection{Reward Structure}
To find an optimal policy, the reward structure is crucial for a reinforcement learning problem. In the literature, there are some basic discrete reward functions, but adding road track information and deviation of a vehicle from the road center is important for realistic training. The customized reward function for the training procedures is given in Eq. \ref{rew}. The reward is calculated at every control step.

\begin{equation}
R = sp \times (cos(\beta) - |sin(\beta)| - tr_{pos}) \label{rew}    
\end{equation}
$sp$ is the speed of the vehicle in $km/h$ obtained from the resultant value of lateral and longitudinal axes speeds. $\beta$ is the angle in radians and it shows the angular deviation of the vehicle heading direction and road central axis. $tr_{pos}$ is the lateral deviation of the vehicle from the road center in meters. Eq. \ref{rew} states that if the speed of a vehicle is high, resultant reward gets higher; but the agent would have to minimize the result of $|sin(\beta)|$ and $tr_{pos}$ to get a positive reward. In the reward function, the speed ($sp$) is multiplied with a difference of the road deviation angle’s cosine and sine components to attain members of vehicle's speed in lateral and longitudinal motion. Another variable in Eq. \ref{rew} is $tr_{pos}$ which is the distance of the agent from the central axis in meters. Minus sign of $tr_{pos}$ multiplied with resultant speed of the vehicle makes the negative reward component of total reward calculation. The agent should learn to stay at the center of the road as much as possible because of this negative reward.
\iffalse
\smallskip

$-20$ penalty is given to the agent if it crashes, gets damaged, gets out of the road or if its speed is too low. In all these situations, the agent gets a constant penalty, which makes the penalization process equal for all terminations.
\fi
\subsection{Hyper-parameters} 
\begin{table}[H]
\caption{Simulation Parameters} 
\vspace*{-2mm}
\centering    
\begin{tabular}{l c}   
\hline  \hline \vspace*{-2mm}&\\
Value Learning Rate, $lr_{value}$ & 0.0005\\
Value Learning Rate, $lr_{policy}$ & 0.0001\\
Gamma, $\gamma$ & 0.995   \\ 
Theta, $\theta$ & 0.15   \\ 
Tau, $\tau$ & 0.001   \\ 
Alpha, $\alpha$ & 0.2  \\ 
Batch Size & 64   \\ 
Buffer Size & 100000  \\ 
State (Input) Size & 23  \\ 
\hline                                              
\end{tabular}
\label{table:training}                                
\end{table}

The training is initialized with predetermined hyper-parameters which are shown in Table \ref{table:training}. The same hyper-parameters are used in all training cases in this paper for the SAC agent. The state vector consist of $23$ different inputs. $19$ of these inputs come from the laser sensors. The other inputs are related to vehicle position on the track and the current velocity of the vehicle in both X and Y directions.

\section{Results}
\label{sec:results}
 At first, the non-curriculum training procedures are carried out on 9 base tracks (3 weather scenarios for 3 road types) as illustrated in Table \ref{table:tra_combs}. In baseline trainings, the iteration limits are set according to track difficulty levels. The straight road scenarios are trained for $75000$ iterations, U-Turn scenarios are trained for $50000$ iterations and the Circuit Race Track scenarios are trained for $200000$ iterations. After the baseline non-curriculum trainings, a $200000$ iteration limit is decided for the training of  curriculum scenarios. In each of 5 different curriculum learning scenarios, 3 different road and weather combinations are used. The first step of all of these curriculum scenarios is one of the baseline trainings that are conducted.
 
\begin{table}[H]
\caption{Baseline Training Results}
\begin{tabular}{|c|c|c|c|}
\hline
\begin{tabular}[c]{@{}c@{}}Road Type /\\ Weather Condition\end{tabular} & Straight Road              & U-Turn Road              &  Race Track             \\ \hline
Clear  &$138,496$   &$32,692$   &$95,237$   \\ \hline
Rainy  &$63,155$    &$23,571$   &$74,476$   \\ \hline
Snowy  &$39,350$    &$16,597$   &$70,742$   \\ \hline
\end{tabular}
\label{table:baseline}
\end{table}

The baseline training results can be seen in Table \ref{table:baseline}. As expected, the lowest rewards for all three road types are recorded on snowy environment and the highest reward is achieved in clear weather.
\smallskip
%\subsection{Straight (C), U-Turn (C), Race Track (C) - (Scenario 1)}

\begin{table}[H]
\vspace*{-2mm}
\centering    
\begin{tabular}{l c}   
\hline  \hline \vspace*{-2mm}&\\
Straight - Clear & It = $75000$  | Rew = $138,496 \pm54\% $ \\
U-Turn - Clear   & It = $25000$  | Rew = $42,429 \pm38\% $\\
Circuit Race Track - Clear  & It = $100000$ | Rew = $843,775 \pm18\% $\\
\hline                                              
\end{tabular} 
\caption{Results of Scenario 1} 
\label{table:purple}                                
\end{table}

%\subsection{U-Turn (R), Race Track (R), Race Track (C) - (Scenario 2)}
\begin{table}[H]
\vspace*{-2mm}
\centering    
\begin{tabular}{l c}   
\hline  \hline \vspace*{-2mm}&\\
U-Turn - Rainy  & It = $50000$  | Rew = $23,571 \pm43\% $\\
Circuit Race Track - Rainy  & It = $75000$  | Rew = $63,015 \pm43\%$\\
Circuit Race Track - Clear   & It = $75000$ | Rew = $144,426 \pm24\%$\\
\hline                                              
\end{tabular}
\caption{Results of Scenario 2} 
\label{table:orange}    
%s\caption{\kXX{Why are there no captions for these tables? Leave no table without a caption!}}
\end{table}

%\subsection{Straight (R), Race Track (R), Race Track (S) - (Scenario 3)}
\begin{table}[H]
\vspace*{-2mm}
\centering    
\begin{tabular}{l c}   
\hline  \hline \vspace*{-2mm}&\\
Straight - Rainy & It = $75000$  | Rew = $63,109 \pm43\%$\\
Circuit Race Track - Rainy & It = $75000$  | Rew = $83,222 \pm56\%$\\
Circuit Race Track - Snowy & It = $50000$ | Rew = $57,869 \pm27\%$\\
\hline                                              
\end{tabular}
\caption{Results of Scenario 3} 
\label{table:green}    
\end{table}
 
%\subsection{Straight (C), Race Track (R), Race Track (S) - (Scenario 4)}
\begin{table}[H]
\vspace*{-2mm}
\centering    
\begin{tabular}{l c}   
\hline  \hline \vspace*{-2mm}&\\
Straight - Clear & It = $75000$  | Rew = $138,496 \pm45\%$\\
Circuit Race Track - Rainy & It = $75000$  | Rew = $1,117,065 \pm35\%$\\
Circuit Race Track - Snowy & It = $50000$ | Rew = $1,248,453 \pm29\%$\\
\hline                                              
\end{tabular}
\caption{Results of Scenario 4} 
\label{table:blue}      
\end{table}

%\subsection{Straight (C), Straight (S), Race Track (S) - (Scenario 5)}
\begin{table}[H]
\vspace*{-2mm}
\centering    
\begin{tabular}{l c}   
\hline  \hline \vspace*{-2mm}&\\
Straight - Clear & It = $75000$  | Rew = $138,496 \pm47\%$\\
Straight - Snowy & It = $25000$  | Rew = $143,405 \pm63\%$\\
Circuit Race Track - Snowy & It = $100000$ | Rew = $597,622 \pm19\%$\\
\hline                                              
\end{tabular}
\caption{Results of Scenario 5}
\label{table:red}      
\end{table}

\begin{figure}[H]
    \centering
    \includegraphics[width=0.475\textwidth]{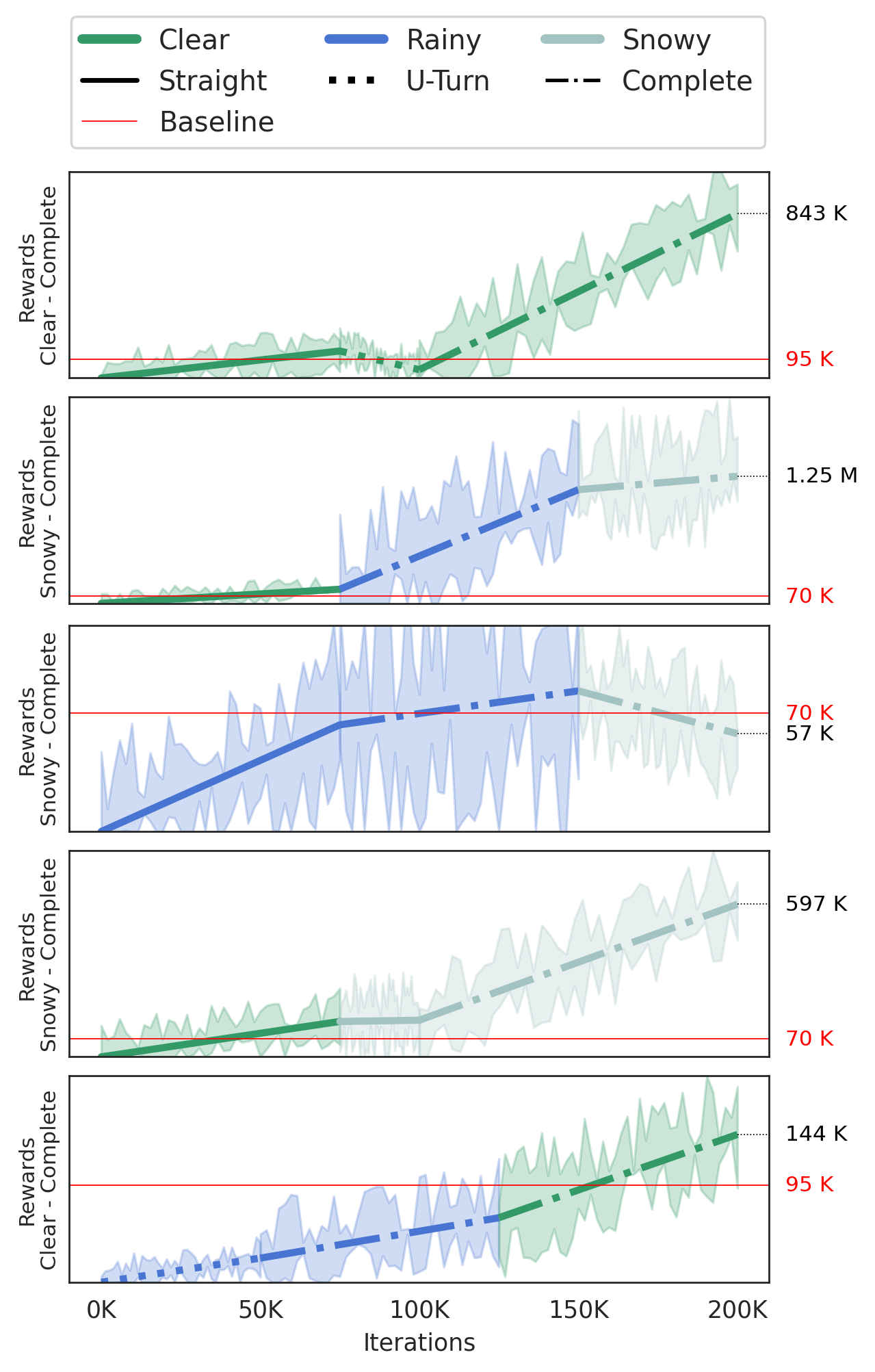}
    \caption{Comparative curriculum learning results. The performances of the models trained with 3 different seeds are shown. Y-Axis represents the rewards, X-Axis represents the environment iterations.}
    \label{Comparative Curriculum Learning Results}
\end{figure}

In the Figure \ref{Comparative Curriculum Learning Results}, it is possible to see the complete picture of the curriculum learning results.

 There are 5 rows in the figure and each one of them represents a separate curriculum scenario. Each training phase executed with 3 different pre-defined seeds for the sake of stability. The left side of the figure shows the ending point of the curriculum scenario and right side shows the baseline and curriculum training results. The red lines in the right side shows the baseline training results for that particular curriculum scenario and the black one shows the curriculum training results. Each weather condition is represented with a different color. The clear weather is shown with green, the rainy weather is shown with blue and the snowy weather is shown with celeste color. In the same fashion, all road types are shown with different markers and they can be seen in the upper part of the Figure  \ref{Comparative Curriculum Learning Results}.

\iffalse
Up until this point, it is shown that the curriculum learning significantly boosts the performance of the trained RL agents and aids them to learn the underlying dynamics of the different weather and road combinations. In addition to this advantage, it is observed that curriculum learning also decreases the training times significantly. For baseline Circuit Race Track with rainy weather, the best reward achieved after $200,000$ iterations is $74,476$. In the Scenario 4 in Table \ref{table:blue}, after $150,000$ iterations, the maximum reward achieved is $1,117,065$. Not only curriculum learning significantly improved the performance of the agent, it also decreased the training time by $\%25$. This gain in sample complexity is extremely valuable in autonomous driving applications.
\smallskip
\fi

\section{Discussions}

The best results are achieved in a curriculum scenario where the best performing non-curriculum agent used directly in the highest complexity environment. In nearly all of the results, the curriculum learning yielded greater results compared with the non-curriculum results except one case. Using a bad performing non-curriculum agent on a curriculum learning scenario severely hurt the training process and decreased the reward compared with the non-curriculum scenario. This shows that it is crucial to use a good baseline agent in all curricula to achieve better results. The results are affected more by the weather changes than the road complexity. 

\smallskip

It might also be helpful to use that insight when deploying an automated curriculum algorithm for decreasing the total time of training. The algorithm that seeks for the optimal transition of the environment variable which has the most impact on reward (snowy to rainy e.g.) would discover optimal curricula faster.

\iffalse
In Scenario 3, a relatively bad baseline train, which is obtained on a higher difficulty environment, is used and the effects of it on the curriculum process are observed. The baseline training reward is $70,742$ for this case, however after the curriculum training process, the reward decreased to $57,869$ which is shown in Table \ref{table:green}. It shows that using a poorly trained baseline agent in the curriculum process has negatively effected the whole training process. 
\fi
%%%%%%%%%%%%%%%%%%%%%%%%%%%%%%%%%%%%%%%%%%%%%%%%%%%%%%%%%%%%%%%%%%%%%%%%%%%%%%%%%%%%%%%%%%%%%%%%%%%%%%%
\section{Conclusion}
In this work, it is showed that training a deep reinforcement learning agent with curriculum learning strategy increases the performance and decreases the overall training time for an autonomous driving agent trained on different weather and road conditions. These results are concluded by developing a structured reinforcement learning system with different road types and weather conditions. The curriculum scenarios on this work consist of different road geometries and weather combinations. It is illustrated that training an RL agent on the relatively simple environment then continuing the the training process of this agent in a more complex environment resulted in a performance boost. During the training process, all parameters kept constant in order to make sure the validity of the experiments. The experimental results demonstrated that an RL agent trained with a curriculum learning structure performed significantly better than an RL agent trained from scratch without a curriculum approach.

%%%%%%%%%%%%%%%%%%%%%%%%%%%%%%%%%%%%%%%%%%%%%%%%%%%%%%%%%%%%%%%%%%%%%%%%%%%%%%%%%%%%%%%%%%%%%%%%%%%%%%%
\section{Acknowledgements}
This work is supported by Istanbul Technical University BAP Grant NO: MOA-2019-42321 and Eatron Technologies, presented at the Workshop on Autonomy at Scale (WS-52), IV2021.

%\addtolength{\textheight}{-1cm}   % This command serves to balance the column lengths

%%%%%%%%%%%%%%%%%%%%%%%%%%%%%%%%%%%%%%%%%%%%%%%%%%%%%%%%%%%%%%%%%%%%%%%%%%%%%%%%

%%%%%%%%%%%%%%%%%%%%%%%%%%%%%%%%%%%%%%%%%%%%%%%%%%%%%%%%%%%%%%%%%%%%%%%%%%%%%%%%


%%%%%%%%%%%%%%%%%%%%%%%%%%%%%%%%%%%%%%%%%%%%%%%%%%%%%%%%%%%%%%%%%%%%%%%%%%%%%%%%

\bibliography{references}
\bibliographystyle{ieeetr}

\end{document}